\documentclass[conference]{IEEEtran}
\IEEEoverridecommandlockouts
\usepackage{cite}
\usepackage{amsmath,amssymb,amsfonts}
\usepackage{algorithmic}
\usepackage{graphicx}
\usepackage{textcomp}
\usepackage{xcolor}
\usepackage{url}
\usepackage{flushend}
\usepackage{hyperref}
\hypersetup{draft}

\usepackage{resizegather}
\def\BibTeX{{\rm B\kern-.05em{\sc i\kern-.025em b}\kern-.08em
    T\kern-.1667em\lower.7ex\hbox{E}\kern-.125emX}}
\begin{document}

\title{Ancient Coin Classification Using Graph Transduction Games}

\author{
\IEEEauthorblockN{Sinem Aslan\textsuperscript{1,2,4}, Sebastiano Vascon\textsuperscript{1,2,3}, Marcello Pelillo\textsuperscript{1,2,3}}
\IEEEauthorblockA{\textsuperscript{1}\textit{European Centre for Living Technology (ECLT), Ca\textquotesingle \hspace{0.1cm}Foscari University of Venice}, Venice, Italy \\
\textsuperscript{2}\textit{Center for Cultural Heritage Technology (CCHT), Ca\textquotesingle\hspace{0.1cm}Foscari University of Venice}, Venice, Italy\\
\textsuperscript{3}\textit{Department of Environmental Sciences, Informatics and Statistics (DAIS), Ca\textquotesingle\hspace{0.1cm}Foscari University of Venice}, Venice, Italy\\
\textsuperscript{4}\textit{International Computer Institute, Ege University}, \.{I}zmir, Turkey\\
siinem@gmail.com,\{sebastiano.vascon, pelillo\}@unive.it }
}

\maketitle

\begin{abstract}
Recognizing the type of an ancient coin requires theoretical expertise and years of experience in the field of numismatics. Our goal in this work is automatizing this time-consuming and demanding task by a visual classification framework. Specifically, we propose to model ancient coin image classification using Graph Transduction Games (GTG). GTG casts the classification problem as a non-cooperative game where the players (the coin images) decide their strategies (class labels) according to the choices made by the others, which results with a global consensus at the final labeling. Experiments are conducted on the only publicly available dataset which is composed of 180 images of 60 types of Roman coins. We demonstrate that our approach outperforms the literature work on the same dataset with the classification accuracy of 73.6\% and 87.3\% when there are one and two images per class in the training set, respectively.  
\end{abstract}


\section{Introduction}
Ancient coins, that depict cultural, political and military events, natural phenomena, ideologies and portraits of god and emperors are important source of information for historians and archaeologists. Recognizing the type of an ancient coin requires theoretical expertise and years of experience in the field of numismatics. A common way to detect the period of a discovered coin is searching through the manual books where ancient coins are indexed \cite{crawford1974roman}  which requires a highly time consuming labor. Our goal in this paper is automatizing recognition of Roman coins by employing computer vision and pattern recognition techniques. Automatizing such a manual procedure not only provides faster processing time but also can support historians and archaeologists for a more accurate decision. A visual classification framework for ancient coin recognition can also be used at museums or by individual collectors to organize large collections of coins.

\begin{figure}[t!]
 \centering
  \includegraphics[width=0.8\linewidth]{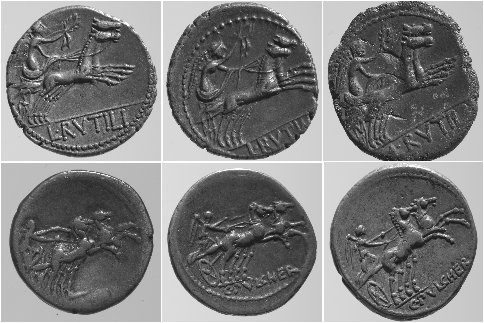}
\label{fig:1}
 \caption{Example images of two classes from the Roman coin dataset \cite{zambanini2012coarse} that is used in this work. \textit{First row:} Images of class 387/1; \textit{Second row:} Images of class 300/1 (listed with Crawford \cite{crawford1974roman} reference number).}
\end{figure}

From the computer vision point of view, classification of ancient coin images is a highly challenging task. One of the difficulties arises from existence of high number of types (i.e. classes) in ancient coins (e.g. Portuguese coins from medieval period and coins from Roman Republic compose over 1500 \cite{salgado2016medieval} and 550
\cite{crawford1974roman} different classes, respectively), while most of the classes include few known specimens as mentioned in \cite{salgado2016medieval,zambanini2014classifying}. Moreover, intra-class variation is large due to local spatial variations arising from missing parts and degradations on the coins, and manual manufacturing of coins by different engravers. Another reason of large intra-class variation is the metallic structure of these coins yields to strong reflection and shading variations so the appearance of the same coin changes significantly under different lighting conditions. Another challenge in ancient coin classification is the typical low inter-class variations due to high global similarity between classes \cite{zambanini2011automatic}. Images from two coin classes are presented in Fig. \ref{fig:1} to demonstrate the challenges of large intra-class and low inter-class variations.

Ancient coin classification can be accomplished by adopting one of the following approaches for classifiers \cite{boiman2008defense}: (i) \textit{learning-based classifiers}, where the parameters of the classifier (e.g. Deep Neural Networks, SVM, Random Forests, etc.) are learned from data in an intensive training phase. (ii) \textit{non-parametric classifiers}, where the classification decision is directly based on data without pursuing any training phase (e.g. Nearest Neighbor based classifier). Although the first group proved to be superior to the second one, they require extraction of highly discriminative features (possibly from abundant training data) for robust classification. Moreover, pursuing such a time consuming training phase can be impractical for handling dynamic databases where new classes are included continuously. 

In this paper, we adopt a non-parametric classifier for ancient coin classification, which is preferable under existence of aforementioned challenges, i.e. large intra-class and low inter-class variation and lack of abundant training data. We have followed the same approach in \cite{zambanini2011automatic,zambanini2012coarse}, i.e. our non-parametric classifier uses a dissimilarity measurement derived from costs of dense matching of SIFT features. 
Similar to \cite{zambanini2011automatic,zambanini2012coarse}, for dense feature matching we use SIFT flow \cite{liu2011sift}, a flow estimation technique developed for image alignment. SIFT flow preserves discontinuity so allows matching objects that locate at different parts of image. This property of SIFT flow makes it well suited for coin images \cite{zambanini2011automatic}, i.e. it helps to deal with large intra-class variation since images from the same class has similar spatial arrangement of features. Additionally, defining similarity between two coin images based on local matches between them helps to deal with low inter-class variation, since two classes mostly differ from each other in variations at local regions. 

Differently from \cite{zambanini2011automatic,zambanini2012coarse}, in this work we do not use a greedy Nearest Neighbor (\textit{NN}) based classifier where a query image is labeled with the class of its nearest (most similar) image in the dataset. Instead, we use a semi-supervised learning approach, namely \textit{Graph Transduction Game} (\textit{GTG}) \cite{DBLP:journals/neco/ErdemP12}, for ancient coin classification. The GTG casts the classification problem as a non-cooperative game where the players (the coin images) decide their strategies (class labels) according to the choices made by the others, which results with a global consensus at the final labeling. Experimenting on a small-scale ancient coin dataset having the aforementioned classification challenges, we show that the notion of label consistency \cite{hummel1983foundations} provided by GTG brings significant performance gain over the conventional NN-based classifier for this challenging problem. 

\section{Previous works}
One of the main problems of ancient coin image analysis that is addressed in the literature is \textit{coin identification} where the goal is recognizing a specific coin instance instead of a coin type \cite{huber2011identification, kampel2009image}. 
This type of application finds usage at identification  of  stolen  coins. Most of the other works have focused on \textit{coin type recognition} (or \textit{coin classification}) which has found a wider range of practical usage. 
A number of works \cite{kampel2008recognizing, zambanini2011automatic, zambanini2012coarse, zambanini2014classifying} employed NN-based classifier where the class of a query coin image is assigned with the label of its most similar one in the training set. Among these, \cite{kampel2008recognizing} defined coin similarity by number of matched SIFT features that were detected sparsely on the images, while \cite{zambanini2011automatic,zambanini2012coarse} employed dense matching costs of SIFT flow as dissimilarity metric. In \cite{zambanini2014classifying}, the authors used densely computed illumination-invariant LIDRIC features and fusing several similarity scores that point out the matching quality they employed an overall similarity score. High performance results are reported in these works although the employed datasets were quite small-scale, i.e. classification accuracies of 90\% \cite{kampel2008recognizing} and 82\% \cite{zambanini2012coarse} are obtained for the datasets with 390 images of 3 classes and 180 images of 60 classes, respectively.  

Other works employed learning-based classifiers. Earlier attempts \cite{anwar2015coarse,arandjelovic2010automatic} relied on Bag of Visual Words based representation of local image features where a visual dictionary is learned from a training set and classification is achieved with SVM in \cite{anwar2015coarse} and GMM in \cite{arandjelovic2010automatic}. Recently, Schlag and Arandjelovic proposed to use a deep convolutional neural network for Roman coin classification in \cite{schlag2017ancient}. They accomplish training with a large set of images, i.e. around 20K images of 83 classes, and reported around 83\% accuracy on 10k images. 

A significant obstacle at employing learning-based classifiers for this particular research problem is deficiency of publicly available datasets. A number of works employed datasets of Sassanian dynasty coins \cite{parsa2017coarse}, some others focused on medieval coins \cite{salgado2016medieval}, and most of them have worked on coins of the Roman Republic \cite{anwar2015coarse,arandjelovic2010automatic,schlag2017ancient,zambanini2011automatic,zambanini2012coarse,zambanini2014classifying}. However, the only publicly available ancient coin dataset is published by Zambanini and Kampel which is composed of 180 images of 60 Roman coin classes \cite{zambanini2012coarse} which we experimented on in this work. 


\section{Graph Transduction Game}
The Graph Transduction Game (GTG) \cite{DBLP:journals/neco/ErdemP12}, is a semi-supervised learning method which has recently found a renewed interest and successfully applied in different contexts, e.g. bioinformatics \cite{DBLP:journals/jmiv/vascon2018} and label augmentation problem \cite{elezi2018transductive}.
The GTG casts the problem in terms of a non-cooperative multiplayer game, in which the objects (or images of a dataset) are the players while the possible strategies are the class labels. 
The idea is, randomly taking two players, they both choose a strategy with a certain probability and receive a payoff which is proportional to the agreement of the chosen strategies (labels). Being a non-cooperative game is in their own interest to maximize their payoff, hence choosing the labels with the higher agreement. Then, the game is played until all the objects have chosen a strategy (label) and none of them would like to change their membership hypothesis. This particular condition is known as \emph{Nash Equilibria} \cite{Nash1951}. Once the game reaches an equilibrium, every player plays its best strategy which correspond to a consistent labeling \cite{hummel1983foundations,miller1991copositive}.
A peculiarity of GTG is that the consistency is a global property which is not related to a single player but achieved for all of the players. 

For the sake of completeness we recap some basic concepts on game-theory in here. Given a set of players $I = \{1, \dots, n\}$ (i.e. images of our dataset) and a set of possible pure strategies $S = \{1, \dots, m\}$ (the set of labels):
\begin{enumerate}
\item \emph{mixed strategy}: a mixed strategy $x_i$ is a probability distribution over the possible strategies for player $i$. Then, $x_i \in \Delta^m$, where 
$\Delta^m = \left\lbrace \sum_{h = 1}^{m} x_i(h) = 1, x_i(h) \geq 0, \ h = \{1, \dots, m\} \right\rbrace$
is the standard $m$-dimensional simplex and $x_i(h)$ is the probability of player $i$ to choose the pure strategy $h$.
\item \emph{strategy space}: 
it corresponds to the set of all mixed strategies of the players
$x = \left\lbrace x_1, \dots, x_n \right\rbrace$, which is represented as a stochastic matrix of size $n \times m$. 
The starting point of the game is defined by a proper setting of $x$.
\item \emph{utility function}: it is responsible for computing the gain obtained by the $i$-th player when it chooses a mixed strategy $x_i$. In particular $u: x_i \rightarrow \mathbb{R}_{\geq 0}$.
\end{enumerate}

In this context, the players are separated into \emph{labeled} ($\mathcal{L} \subset I$) and \emph{unlabeled} ($\mathcal{U} \subset I$) sets\footnote{In terms of standard learning algorithm, the set of labeled players correspond to the training set while the unlabeled ones to the test set.}. The strategy space $x$ is initialized 
in two different ways based on the fact that an object is \emph{labeled} or \emph{unlabeled}.
 A one-hot vector is assigned to each of the labeled objects, since their labels are known: 
\begin{equation}\label{eq:Xprior}
x_i(h)=
\begin{cases}
	1, & \text{if } i \text{ has label } h \\
    0, & \text{ otherwise}.
\end{cases}
\end{equation}

whereas, since no prior knowledge is available for the unlabeled objects, the same probability of all labels is assigned to them:
\begin{equation}
x_i(h) = \frac{1}{m} \quad \text{ } \forall h \in S\label{eq:Xnoprior}
\end{equation}

\paragraph{Payoff function} 
The payoff function reflects the likelihood for a player (object) to choose a particular strategy (label), considering the similarities between labeled and unlabeled players. 
It provides that more similar players are more likely to influence each other in choosing one of the possible strategies (labels). 

Formally, given a player $i$ and a strategy $h$ the utility function is as follows: 
\begin{eqnarray}
\mathcal{N}_i &=& k \text{-nearest neighbors of player } i \label{eqn:neighbors}\\
u_i(h)&=&\sum_{j \in \mathcal{U}_i}{(A_{ij}x_j)_h}+\sum_{\gamma=1}^{m}{\sum_{j \in \mathcal{L}_i}{A_{ij}(h,\gamma)}}  \label{eqn:payoff_function_single}
\\
u_i(x)&=&\sum_{j \in \mathcal{U}_i}{x_i^TA_{ij}x_j}+\sum_{\gamma=1}^{m}{\sum_{j \in \mathcal{L}_i}{x_i^T(A_{ij})_\gamma}}
\label{eqn:payoff_function}
\end{eqnarray}

where $\mathcal{L}_i \subseteq \mathcal{N}_i$ and $\mathcal{U}_i \subseteq \mathcal{N}_i$ are the labeled and unlabeled nearest neighbors of $i$, respectively. Here, $u_i(h)$ and $u_i(x)$ are the payoffs received by player $i$ while it uses the strategy $h$ and plays the mixed strategy $x_i$, respectively.
The matrix $A_{ij} \in \mathbb{R}^{m \times m}$ is the \emph{partial payoff matrix} between players $i$ and $j$, which is computed as $A_{ij} = I_m \times \omega(i,j)$ \cite{DBLP:journals/neco/ErdemP12},
where $\omega(i,j)$ is the similarity between player $i$ and $j$ and $I_m$ is the identity matrix of size $m \times m$.

\paragraph{Players similarity}
Once the features are extracted for players (objects or images) $i$ and $j$, similarity between them can be computed by Eq. \ref{eqn:similarity}, where $d(f_i,f_j)$ denotes the distance between features $f_i$ and $f_j$ and $\sigma_{i}$ is the distance between $i$ and its $7$-nearest-neighbors \cite{DBLP:conf/nips/Zelnik-ManorP04}. 
\begin{equation}\label{eqn:similarity}
\omega(i,j)=exp\left\lbrace-\frac{d(f_i,f_j)}{\sigma_{i}\text{ }\sigma_{j}}\right\rbrace
\end{equation}
\paragraph{Finding Nash Equilibria}
In order to find a Nash Equilibria of the game 
we used a result, named as Replicator Dynamics (RD) \cite{smith1982evolution}, from Evolutionary Game Theory\cite{weibull1997evolutionary}. 
The RD are dynamical systems that mimic a Darwinian selection process on a set of strategies for each player. The underlying idea is it 
favors the fittest strategies 
for their survival while the others 
become extinct. 

More formally, the RD are defined as follows: 
\begin{equation}\label{eq:rd}
x_i(h)^{t + 1} = x_i(h)^{t}\frac{u_i(h)^t}{u_i(x^t)}
\end{equation}

where $x_i(h)^{t}$ is the probability of strategy $h$ at time $t$ for player $i$ (see Eq. \ref{eqn:payoff_function_single}) and $u_i(x^t)$ is the expected payoff of the entire mixed strategy (see Eq. \ref{eqn:payoff_function}). The Eq. \ref{eq:rd} is iterated until convergence\footnote{Convergence criteria: \emph{i)} the distance between two successive steps is $||x^{t+1}-x^t||_2 \leq \varepsilon$ or \emph{ii)} a certain amount of iterations is reached, i.e. typically 20 iterations are sufficient.} (See \cite{DBLP:journals/jmiv/Pelillo97} for a detailed analysis). Once the convergence of Eq.\ref{eq:rd} is reached, we simply get the index of the maximum value in the $i$-th row of $x$ in order to label the $i$-th object.
\section{Ancient Coin Classification using Graph Transduction Game (GTG)}
By considering the training set of coin images as the labelled players, GTG can be applied for ancient coin classification problem to estimate the labels of the test set images, i.e. unlabelled players. We list the steps that we have employed for the application of GTG for ancient coin classification as follows:
\paragraph{\textbf{Feature extraction}} We compute two type of features on the images: (i) In order to analyze local similarities, we compute 128-dimensional SIFT features in the local neighborhood of every image pixel that results with a tensor named as \emph{SIFT-image} \cite{liu2011sift}; (ii) In order to analyze global similarities between images we compute CNN features. Specifically, since our dataset is quite small, which makes a CNN training unfeasible, we apply transfer learning by using a CNN architecture pre-trained on ImageNet. 
Finally, for each input image we get its feature from the output of the last fully-connected layer of the CNN.

\paragraph{\textbf{Initialization of the strategy space}} 
Since no other knowledge on the problem exists, but only the distinction between labeled and unlabeled sets, the strategy space is initialized using Eq.\ref{eq:Xprior} and Eq. \ref{eq:Xnoprior}.

\paragraph{\textbf{Computation of similarity between objects}} A correct choice of computation for the similarity between images is important to avoid a failure at label estimation. 
We employ different schemes of similarity computation regarding to the extracted feature types: 
 
\textit{i. Similarity between local features of images:} It is demonstrated in \cite{zambanini2011automatic} that matching scores of SIFT flow technique are powerful dissimilarity metric for ancient coin classification. 
In SIFT flow, 
SIFT-images are matched along the flow vectors and optimal correspondences are found by minimizing an energy function ($E(w)$ in \cite{liu2011sift}) using dual-layer belief  propagation \cite{liu2011sift}. 
Since runtime of such optimization 
scales up with the image size, authors of \cite{liu2011sift} proposed to employ coarse-to-fine search which results with faster computation and better performance of matching. Similar to \cite{zambanini2012coarse}, in this work we used the minimum energy value, say $E^*_{i,j}$, (to which SIFT Flow algorithm converges at the finest level of the coarse-to-fine search) as a dissimilarity metric between image $i$ and $j$, i.e. we used $d(f_i,f_j) = E^*_{i,j}$ in Eq. \ref{eqn:similarity}.  

\textit{ii. Similarity between global descriptions of images:} Following the general trend \cite{elezi2018transductive,DBLP:journals/neco/ErdemP12}, we used Euclidean distance, i.e. $d(f_i,f_j)=||f_i-f_j||_2$ in Eq. \ref{eqn:similarity}, to compute similarity between the CNN features.

\paragraph{\textbf{Execution of transduction game}} Giving the similarities 
to the GTG, it starts to play the game between players, i.e. images, until convergence. We get the final probabilities of strategies, i.e. labels, for the unlabeled objects at the output and we assign the object with the strategy that could get the highest maximum probability.

\section{Experiments}

\paragraph{Dataset}    
We experimented on the only published\footnote{\url{http://cvl.tuwien.ac.at/research/cvl-databases/coin-image-dataset/}} 
ancient coin dataset \cite{zambanini2012coarse} which is acquired at Coin Cabinet of the Museum of Fine Arts in Vienna, Austria. The dataset is composed of 180 images (reverse sides of the coins that includes motifs and legends) of 60 classes with 3 images in each class. Images are resized to $150 \times 150$ pixels as in \cite{zambanini2012coarse}.
\paragraph{Experimental setup} Since we have experimented on the same dataset, we followed the same experimental setting with \cite{zambanini2012coarse} to make a fair comparison of techniques. 
In \cite{zambanini2012coarse}, accepting one of the coins as a query image (or test image), the remaining one or two images per class are used to create the training set. At each classification run, nearest neighbor of the query image is searched in the training set. This procedure leads to 180 and 360 classification runs when two and one training images per class is used. When the training set is created by two images per class, the nearest neighbor search is performed through accumulated dissimilarity values of each training set image over classes.




Adopting the same experimental setting in our approach, we create a dissimilarity matrix with the entries computed as in \cite{zambanini2012coarse}, i.e. as mentioned in Section IV.c. Then we symmetrize it (by getting maximum of entries around diagonal) before giving input to the GTG algorithm. Additionally, at each iteration we substitute the test image and training images as unlabeled object and labeled objects, respectively to be used in GTG and we get the class label of the unlabeled object in the output. In all experiments, the parameter $k$ of the neighboring set $\mathcal{N}_i$ in Eq. \ref{eqn:neighbors} is set to 2. 

\paragraph{Performance evaluation} We performed GTG by employing two feature types and with the corresponding dissimilarity metrics as explained in Section IV. In the first experiment, we compute off-the-shelf CNN features by DenseNet-201 which is one of the state-of-the-art CNN architectures where we use the Euclidean distance metric to measure the dissimilarity between the features. In the second experiment, by employing densely computed local SIFT features we use matching costs of SIFT flow as dissimilarity measure. The performance results of these experiments and comparison with the state-of-the-art work on the same dataset \cite{zambanini2012coarse} are given in Table \ref{tab:1}. 

It can be seen in Table \ref{tab:1} that the lowest performance results for both training set sizes are obtained when we use the CNN features. This is an expected outcome, because CNN features provide a global description of images and a high global similarity exists between different classes in this coin dataset. We could outperform \cite{zambanini2012coarse} that employs a NN-based classifier, by using the GTG for ancient coin classification by 73.6\% and 87.2\% classification accuracy when the training set is constructed from one and two images per class, respectively. We additionally checked the performance of conventional NN-based classifier which does not adopt the accumulation of class-wise dissimilarities (that were adopted at \cite{zambanini2012coarse}), when there are two images per class in the training set. In that case, we got 81.67
\% accuracy which was slightly lower than the reported performance (83.3\%) in \cite{zambanini2012coarse}. 

\begin{table*}[tb]
\centering
\caption{Classification results}
\label{tab:1}
\begin{tabular}{p{5.5cm}p{2cm}p{1.8cm}p{2cm}p{1.8cm}}
\hline

 & \multicolumn{2}{c}{Training set: 1 image per class} & \multicolumn{2}{c}{Training set: 2 images per class} \\
\cline{2-5}
Technique & Correct classifications & Classification accuracy & Correct classifications & Classification accuracy \\
\hline
CNN features + Euclidean distance + GTG & 188 / 360 & 52.2\% & 113 / 180 & 62.8\% \\
Dense SIFT + Matching cost + NN \cite{zambanini2012coarse} & 257 / 360 & 71.4\% & 150 / 180 & 83.3\% \\
\textbf{Dense SIFT + Matching cost + GTG}  & \textbf{265 / 360} & \textbf{73.6\%} & \textbf{157 / 180} & \textbf{87.2\%} \\
\hline
\end{tabular}
\end{table*}

\begin{figure}[t!]
 \centering
  \includegraphics[width=0.8\linewidth]{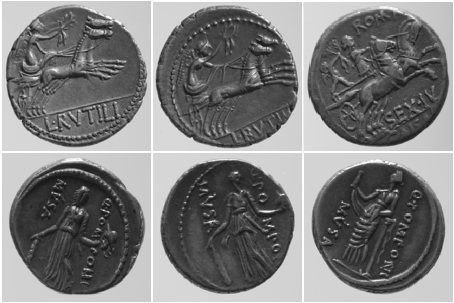}
 \caption{Two selected misclassifications of the proposed approach based on GTG. \textit{First column:} test image; \textit{Second column:} another image from the same class; \textit{Third column:} image of selected class by the proposed scheme.}
 \label{fig:2}
\end{figure}

In Fig. \ref{fig:2}, we present two misclassifications of the proposed approach. It can be seen that the misclassifications are mostly due to low variability between different classes.  

\section{Conclusion}
In this paper, we studied the ancient coin classification problem using Graph Transduction Games (GTG) which adopts the approach of non-parametric classifier. 
The GTG is a game-theoretic semi-supervised learning algorithm, grounded on the notion of label consistency, in which the final labeling of the objects is achieved by reaching an equilibrium condition between all labeling hypothesis.
Our experimental results show that GTG works better for the problem of ancient coin classification, which is a highly complex problem due to large intra-class and low inter-class variations, compared to conventional nearest neighbor based non-parametric classifiers that does not consider global agreement at labeling choices of all dataset images.

\section*{Acknowledgment}
The authors would like to thank Sebastian Zambanini, Ismail Elezi, Leulseged Tesfaye Alemu and Alessandro Torcinovich for their invaluable advices, sharing and helps at various technical issues. 

\bibliographystyle{IEEEtranS}
\bibliography{conference_041818}

\end{document}